\documentclass[sigconf,authorversion,nonacm]{acmart}

\AtBeginDocument{%
  }

\copyrightyear{2025}
\acmYear{2025}
\setcopyright{licensedothergov}\acmConference[K-CAP '25]{The 13th International Conference on Knowledge Capture}{December 10--12, 2025}{Dayton, Ohio, USA}
\acmBooktitle{The 13th International Conference on Knowledge Capture (K-CAP '25), December 10--12, 2025, Dayton, Ohio, USA}
\acmPrice{15.00}
\acmDOI{0000000000000}
\acmISBN{0000000000000}

\usepackage{amsmath,amsfonts}
\usepackage[linesnumbered,ruled,vlined]{algorithm2e}
\usepackage{graphicx}
\usepackage{textcomp}
\usepackage{subcaption}
\usepackage{float}
\usepackage{caption}

\usepackage{mathtools}
\usepackage{xcolor}

\definecolor{neuro_color}{rgb}{0.59, 0.45, 0.65}
\definecolor{symbolic_color}{rgb}{0.51, 0.70, 0.40}

\usepackage{xspace}

\begin{document}

%\title[]{Saliency Map-Based Knowledge Discovery for Subclass Identification Incorporating LLM-Based Symbolic Approximations}
%%% Im Titel ist zweimal "-based"
%%% Vorschlag
\title[]{Saliency Map-Guided Knowledge Discovery for Subclass Identification with LLM-Based Symbolic Approximations}


\author{Tim Bohne}
\email{tim.bohne@dfki.de}
\affiliation{%
  \institution{German Research Center for\\ Artificial Intelligence (DFKI)}
  \streetaddress{Hamburger Straße 24}
  \city{Osnabrück}
  \country{Germany}
}

\author{Anne-Kathrin Patricia Windler}
\email{patricia.windler@dfki.de}
\affiliation{%
  \institution{German Research Center for\\ Artificial Intelligence (DFKI)}
  \streetaddress{Hamburger Straße 24}
  \city{Osnabrück}
  \country{Germany}
}

\author{Martin Atzmueller}
\email{martin.atzmueller@uos.de}
\affiliation{%
  \institution{Semantic Information Systems, Osnabrück University, and \mbox{DFKI}}
  \streetaddress{Wachsbleiche 27}
  \city{Osnabrück}
  \country{Germany}
}

\renewcommand{\shortauthors}{Bohne et al.}

\begin{abstract}
This paper proposes a novel neuro-symbolic approach for sensor signal-based knowledge discovery, focusing on identifying latent subclasses in time series classification tasks.
The approach leverages gradient-based saliency maps derived from trained neural networks to guide the discovery process. Multiclass time series classification problems are transformed into
binary classification problems through label subsumption, and classifiers are trained for each of these to yield saliency maps. The input signals, grouped by predicted class, are clustered under three distinct configurations.
The centroids of the final set of clusters are provided as input to an LLM for symbolic approximation and fuzzy knowledge graph matching to discover the underlying subclasses of the original multiclass problem.
Experimental results on well-established time series classification datasets demonstrate the effectiveness of our saliency map-driven method for knowledge discovery, outperforming signal-only baselines in both
clustering and subclass identification.
\end{abstract}




\maketitle

\section{Introduction}

Saliency maps, class activation maps, attribution maps, or, in general, methods highlighting regions of interest (ROI) that were decision-relevant in a deep learning system's
prediction are widely used~\cite{Bohne:2023, Hossain:2024, Mahapatra:2022}. However, they should only be considered as a first step towards knowledge discovery.
Knowledge discovery in this work refers to the process of identifying interpretable, symbolic representations that reveal latent, semantically meaningful structures.
Unlike pattern discovery, which identifies discriminative structures without semantics, knowledge discovery implies the extraction of structured, symbolic information, e.g., in the form of knowledge graph properties.
In~\cite{Bohne:2023}, we leverage gradient-based saliency maps derived from trained neural networks as useful explanation resources, highlighting issues such as overfitting and deviation from expert judgments.
Yet, saliency maps can also support the identification of latent subclasses in classification problems, providing valuable knowledge that is unavailable a priori.
The fundamental premise is that the various subclasses should manifest in distinct saliency maps, which can subsequently be facilitated in clustering,
i.e., subclass discovery via saliency map-aided clustering.
The centroids of the resulting clusters are provided as input to an LLM for symbolic approximation and fuzzy knowledge graph (KG) matching.
This establishes a bidirectional synergy that bridges neural network-based pattern discovery and KG-based knowledge discovery.
According to the taxonomy in~\cite{Kautz:2022}, which categorizes neuro-symbolic systems, our approach belongs to the \emph{Neural | Symbolic} category.
Our core contributions are the demonstration of the potential of saliency maps for pattern and knowledge discovery as well as the
development and evaluation of an effective, novel neuro-symbolic approach for saliency map-driven knowledge discovery, aimed at identifying latent subclasses in time series classification tasks.

The remainder of the paper is structured as follows: Section~\ref{sec:related:work} discusses related work. After that, Section~\ref{sec:method} presents our approach, before
evaluating it in Section~\ref{sec:eval}. Finally, Section~\ref{sec:conclusions} concludes with a summary and
promising directions for future research.

\section{Related Work}\label{sec:related:work}

In~\cite{Bohne:2023}, we anticipated the potential of saliency maps for knowledge discovery, i.e., allowing the identification of latent conceptual structures within the data.
Although saliency map generation for deep learning is mainly applied to image data, there are some evaluations of consistency and robustness
of their explanations when applied to time series classification, e.g.,~\cite{Balestra:2023}. It is about evaluating the saliency map generation techniques themselves, not their
use as means of knowledge discovery.
Most of the literature is either concerned with generating saliency maps, e.g., \emph{Grad-CAM}~\cite{Selvaraju:2017}, \emph{Grad-CAM++}~\cite{Chattopadhyay:2018},
\emph{Score-CAM}~\cite{Wang:2020}, and \emph{LayerCAM}~\cite{Jiang:2021}, or with applying them as explanation tools. A recent case of the latter is the
work of Hossain et al.~\cite{Hossain:2024} in which the authors apply saliency maps generated with \textit{GradCAM} to histopathology images for classification of certain diseases, pinpointing crucial areas for
predictions.

Mahapatra et al.~\cite{Mahapatra:2022} work on self-supervised generalized zero shot learning for medical image classification using interpretable saliency maps.
Moreover, they propose a novel method to generate \textit{GradCAM} saliency maps that are supposed to highlight diseased regions with greater accuracy.
%Their aim is to compensate the lack of classes to be expected, i.e., recognizing seen and unseen classes.
%Along the way, they select representative vectors of disease classes and synthesize features of unseen classes.
They enforce the saliency maps of different classes to be different and ensure that clusters in the space of image and saliency features yield centroids with similar
semantic information.
%The new approach is motivated by the infeasibility of the common approach of a model correlating class attribute vectors (semantic embeddings) and corresponding feature
%representations in their domain.
In particular, they use the saliency maps as an additional constraint for the clustering, as an effective way to incorporate an attention mechanism.
\cite{Mahapatra:2022} shares some common goals with this work, e.g., exploiting information from saliency maps to improve a clustering process.
Nevertheless, they incorporate them into the clustering in a different way, i.e., as part of the optimization criterion.
In contrast, we perform multivariate \textit{Dynamic Time Warping Barycenter Averaging} (DBA) $k$-means clustering~\cite{Petitjean:2011} with one dimension being the signals themselves and the other the corresponding saliency maps.
Additionally, they are not considering the identification of subclasses and also not knowledge discovery.

LLMs have recently been applied to various time series tasks, including reasoning and the alignment of time series to textual descriptions.
For instance, Fons et al.~\cite{Fons:2024} evaluate the performance of five general-purpose LLMs in matching time series to textual descriptions.
They observed a close-to-perfect accuracy of \textit{GPT-4} in intra-dataset matching, where qualitative and quantitative descriptions were given for one dataset.
However, a significantly lower accuracy was observed in cross-dataset matching with qualitative descriptions only.
There has also been work on LLMs specialized on time series reasoning, such as \textit{ChatTS}~\cite{Xie:2024}. A multimodal LLM that facilitates understanding and reasoning on multivariate time series
and the alignment of time series with textual descriptions has been demonstrated to outperform \textit{GPT-4o} in these tasks.
Thus, LLMs are a viable tool for matching time series to textual descriptions.
Similar to ~\cite{Fons:2024}, this work also utilizes these matching capabilities to demonstrate the potential for symbolic approximation.

\section{Method}\label{sec:method}

The saliency map-based knowledge discovery comprises seven high-level steps, which are elaborated in the following subsections:
\begin{enumerate}
    \item Defining an exemplary symbolic knowledge representation, i.e., an ontology (cf. Sec.~\ref{sec:symbolic_kr}).
    \item Selecting suitable (criteria-compliant) multiclass time series classification datasets from the UCR archive~\cite{Dau:2018} (cf. Sec.~\ref{sec:dateset_sel}).
    \item Transforming the multiclass classification problem into a binary classification problem via label subsumption (cf. Sec.~\ref{sec:label_subsumption}).
    \item Training classifiers for binary classification tasks (cf. Sec.~\ref{sec:ann_classification}).
    \item Generating saliency maps based on the trained classifiers and test datasets (cf. Sec.~\ref{sec:cam}).
    \item Clustering of the input signals, grouped by predicted class ($0$, $1$): univariate (signal only), univariate (saliency map only), multivariate (signal + saliency map) (cf. Sec.~\ref{sec:clustering}).
    \item Forwarding of cluster centroids to the LLM for symbolic approximation and fuzzy KG matching (cf. Sec.~\ref{sec:llm_approx}).
\end{enumerate}
Crucially, we regard each dataset as a binary classification problem, although both classes comprise a number of subclasses that we aim to (re)discover.
This is done to demonstrate the feasibility of the method in a setup in which ground truth information is available.
When applying the approach in practice, it begins at step $(4)$, omitting those that are part of the evaluation protocol (steps $1$-$3$).

\subsection{Symbolic Knowledge Representation}
\label{sec:symbolic_kr}

\begin{figure}[H]
    \centering
    \includegraphics[width=0.45\textwidth]{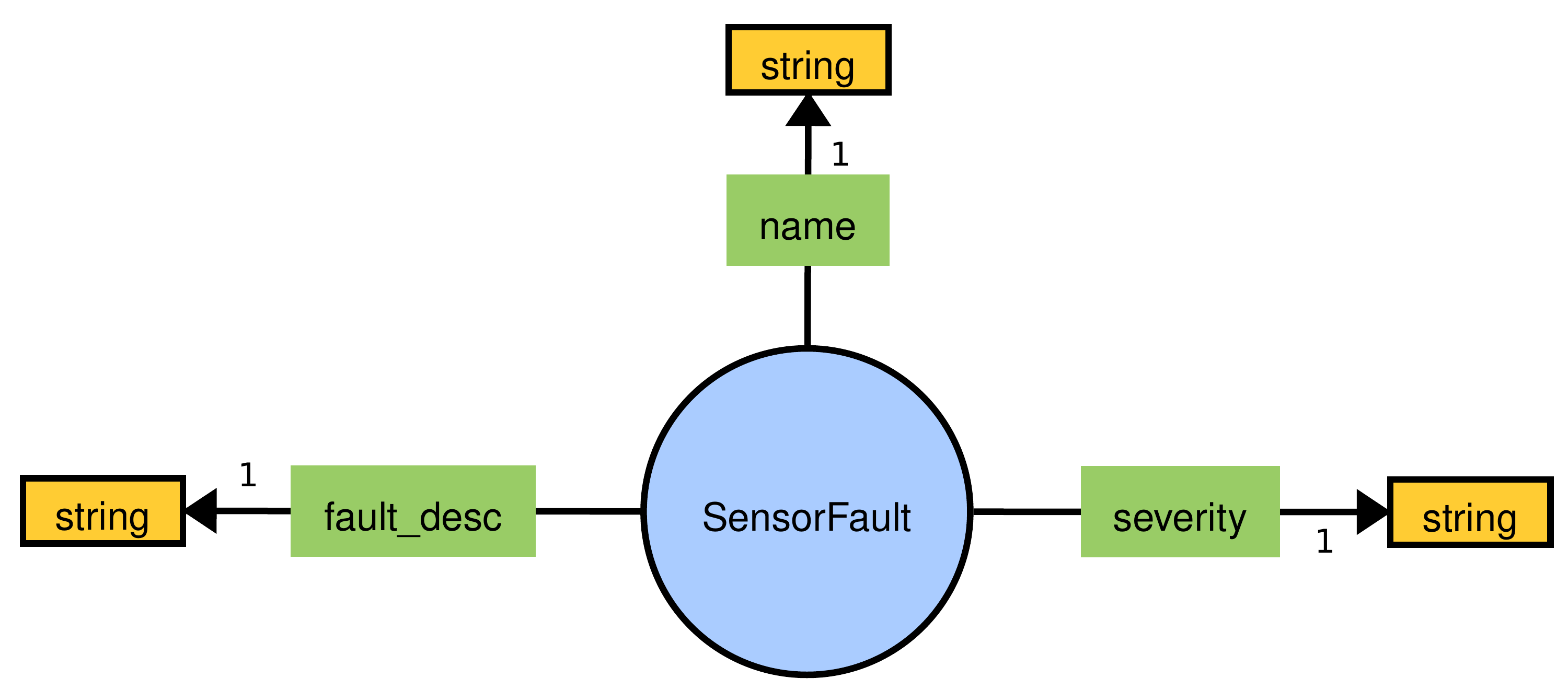}
    \caption[]{Minimalistic Sensor Fault Ontology}
    \label{fig:ontology}
\end{figure}

An aspect of the approach that should not be neglected is the existence of a symbolic knowledge representation, which ultimately turns pattern discovery into a knowledge discovery. Although even the
domain-agnostic characteristics of the symbolic information can be arbitrary for the approach, we will consider general sensor faults as a placeholder for any kind of symbolic description of sensor signals
due to the practical plausibility and the thematic link to our previous
work~\cite{Bohne:2023}.
Maintenance systems often describe signal characteristics linked to problems.
The symbolic class descriptions for UCR datasets are treated as representative sensor fault types.
To capture this domain-specific knowledge, an ontology is defined, and a corresponding KG is constructed for each dataset by populating it with instance data.
The facts are stored in the \textit{Resource Description Framework} (RDF) format and hosted on an \textit{Apache Jena Fuseki} \footnote{https://jena.apache.org/documentation/fuseki2/} server.
Knowledge retrieval as well as KG updates work via predefined SPARQL queries and HTTP requests to the respective endpoints.
The minimalistic ontology used in this paper is shown in Fig.~\ref{fig:ontology}. This could be regarded as an excerpt from a larger ontology contextualizing the sensor faults, e.g., the one used in~\cite{Bohne:2023}.
However, as only this one concept is required to demonstrate the methods in this work, anything superfluous has been omitted.
In the following, each class of each dataset is represented as an instance of \textcolor{darkgray}{\texttt{\textbf{SensorFault}}}. The \textcolor{darkgray}{\texttt{\textbf{name}}} property is always instantiated
with the ground truth label of the corresponding class and \textcolor{darkgray}{\texttt{\textbf{fault\_desc}}} represents the crucial textual representation of the corresponding signal, e.g.,
\textit{``Beginning slightly below zero, it climbs sharply to its main crest, drops into a mid-level dip, rebounds to a smaller hill, and then fades through progressively
weaker ripples into an extended, gently declining negative stretch''}.
Finally, \textcolor{darkgray}{\texttt{\textbf{severity}}} is set to arbitrary values and serves as a placeholder for all kinds of conceivable auxiliary information.
Facilitating this KG, the results of the saliency map-based pattern discovery can be used to assign each discovered subclass to a potentially existing counterpart in the KG,
thereby turning it into an actual symbolic knowledge discovery. If a detected cluster can be matched to some \textcolor{darkgray}{\texttt{\textbf{SensorFault}}}, it is a potential knowledge discovery.
Otherwise, a new subclass not yet part of the KG might have been identified, which should be forwarded to a human expert for verification. There are three cases:
\begin{enumerate}
    \item \textbf{Knowledge discovery}: Identified cluster, matched symbolic description (with uncertainty, cf. Sec.~\ref{sec:eval})
    \item \textbf{Pattern discovery}: Identified cluster, no matching, unknown category to be labeled by a human, i.e., KG extension
    \item \textbf{No discovery}: Identified cluster, no matching, poor quality
\end{enumerate}
The first two are desirable. The last one could be prevented via cluster filtering, e.g., based on ground truth independent metrics presented in Sec.~\ref{sec:eval}.
Given that we transform multiclass classification problems into binary classification problems, the primary goal is to strive for the knowledge discovery case.

\subsection{Dataset Selection}
\label{sec:dateset_sel}

The aim is to transform a multiclass classification problem into a binary classification problem and subsequently (re)discover the remaining subclasses with the help of saliency maps in order to obtain a ground truth evaluable setup.
For this, suitable multiclass datasets from the UCR time series archive~\cite{Dau:2018} are to be used.
For a dataset to be considered suitable, we define certain requirements.
Evidently, the approach demands significantly more than two classes $Y$. Datasets with $|Y| \in [6, 16]$ are considered, in order to avoid
super-classes that exhibit too much variance through a large number of subclasses per class $Y_i$.
To ensure a meaningful evaluation, we only consider datasets that have a large test set with a large number of samples per class -- specifically those containing at least $100$ time series per class $Y_i$.
Datasets comprising time series shorter than $100$ time steps are also excluded from consideration.
Image-to-time-series datasets were disregarded in order to maintain a focus on inherently temporal data.
The classification problem should be reasonably tractable to solve, meaning that it is possible to train a classifier with a decent test accuracy. This is required because an accurate classifier enhances
the probability of obtaining reliable and useful class activation maps. If a classifier produces a large number of misclassifications, it is expected that the class activation maps of the misclassified time series
are less meaningful. Hence, datasets were selected for which classifiers have been trained with a test accuracy of at least $0.6$ according to~\cite{Fawaz:2018}, who studied the performance of different
deep neural network architectures on the $128$ UCR time series datasets.\newline

When filtering the $128$ UCR datasets according to these requirements, the following remain: \textit{InsectWingbeatSound}, \textit{Mallat}, and \textit{UWaveGestureLibraryAll}.
The three selected datasets stem from distinct domains, enabling the evaluation of the approach across varied datasets with differing semantics.
\textit{InsectWingbeatSound}~\cite{Chen:2014} captures the power spectrum of the sound of insects passing through a sensor. Each of the $11$ classes represents a certain type of insect.
It has a training set of size $220$ and a test set of size $1980$, with a time series length of $256$.
\textit{UWaveGestureLibraryAll}~\cite{Liu:2009} consists of recordings of a three-axis accelerometer. The $8$ classes represent gestures performed while holding a handheld accelerometer device.
Each time series has a length of $945$ and consists of the concatenation of the $x$, $y$ and $z$ axes. It has a training set of size $896$ and a test set of size $3582$.
The time series are downsampled to a length of $256$ in the experiments.
\textit{Mallat} is a simulated dataset comprising $8$ classes. The details of its construction are described in~\cite{Mallat:1999}. It has a training set of size $55$, a test set of size $2345$, and the original
length of the time series is $1024$, which is downsampled to $256$ again.\newline

A relevant question is how distinct the classes have to be (inter-class distance) and how similar the samples within one class have to be (intra-class distance) in order for the approach to work.
It is imperative that the samples belonging to a particular subclass are sufficiently homogeneous.
A further intriguing question is whether the best-known classifier accuracy reported in the literature is an appropriate criterion. It could lead to a bias towards favoring easy-to-classify datasets.
The improvement through saliency maps may be particularly significant when the dataset is difficult to classify. On the other hand, a good classifier accuracy should correlate with good saliency maps.
Ultimately, we opted for a classifier accuracy threshold to prevent meaningless saliency maps.

\subsection{Creating Binary Classification Problems via Label Subsumption}
\label{sec:label_subsumption}

As previously discussed, each selected dataset requires label subsumption, i.e., conversion of a multiclass classification problem into a binary classification problem.
This is a straightforward process, in which the class labels $Y$ are subsumed in an alternating fashion:
\[
\text{Y}_i =
\begin{cases}
0 & \text{if } \text{Y}_i \bmod 2 = 0 \\
1 & \text{otherwise}
\end{cases}
\quad \text{for all } i
\]
It is crucial to evenly distribute the original classes among the new binary classes, since subsuming the majority of the classes under one would lead to a strong class imbalance in the binary
classification problem.
Compared to the multiclass classification problem, the binary problem should, in theory, be considerably less challenging.
A question of interest is the importance of aiming for a highly accurate initial binary classifier.
Ideally, this results in enhanced saliency maps, which in turn may facilitate improved clusterings.
The number of subclasses that can be discovered is contingent upon the difficulty of distinguishing the classes and the classifier performance.
Certain subclasses may prove harder to match than others.
It is crucial to note that the objective of this work is not finding the best possible classifier, but rather to demonstrate the feasibility of the principles.
Every single subclass that is identified is gained knowledge.
The rationale behind the challenge of training a binary classifier with high accuracy is that the two classes may contain remarkably distinct signals (i.e., subclasses) based on the particular dataset.
A future work approach could also investigate the selection of the number of classes based on the number of groups, i.e., less than the original
number of classes, but potentially more than two. Then, it would be easier to distinguish the classes and force the model to learn the differences. However, this would also imply a considerable amount of preliminary
analysis.
In general, it is plausible to assign similarly-shaped subclasses to one class in label subsumption. However, then it might suffice to learn some high-level distinctions rather than the nuances, i.e.,
for learning success it might be better to actually assign close subclasses to different classes in the binary problem to enforce learning of nuances.
As previously defined, we assign them randomly (alternating). This may be more aligned with the option that emphasizes learning nuances.
In essence, we regard each dataset as a binary classification problem, although both classes comprise a number of subclasses that we aim to (re)discover via saliency map-based patterns.

\subsection{Binary Time Series Classification}
\label{sec:ann_classification}

The task comes down to binary univariate time series classification.
For a time series $V \in \mathbb{R}^n$, performance was best when standard $z$-normalization was applied to the raw time series data, i.e.,
$V' := \{\frac{x_i - \mu{V}}{\sigma{V}} \thinspace | \thinspace x_i \in V\}$, with mean $\mu{V}$ and standard deviation $\sigma{V}$.
An additional preprocessing step is to downsample each time series to a length of $256$ as a compromise between achieving the highest accuracy and ensuring a reasonable runtime.

Since the main focus of this work is not to propose a superior ANN architecture for binary classification of time series data,
we compared several standard architectures from the \textit{tsai} library and finally selected the best-performing one that allows for a relatively straightforward extraction of saliency maps,
which was the \textit{Explainable Convolutional Neural Network for Multivariate Time Series Classification} (XCM)~\cite{Fauvel:2021}.
An aspect to keep in mind is that the worse the classifier, the worse the saliency maps -- and thus, the higher the chance of the input signals in isolation being more successful in clustering.
Therefore, obtaining high-accuracy binary classifiers remains a concern.
The hyperparameters are always based on several initial test experiments for each dataset.
Table~\ref{tab:binary_res} shows the binary classification results.
The third column refers to the batch size (training / validation), and the fourth to the test accuracy.
The following sections are based on the performance on the test data.\newline
\begin{table}[H]
\centering
\small{
\begin{tabular}{|c|c|c|c|c|}
\hline
\textbf{dataset} & \textbf{architecture} & \textbf{bs} & \textbf{epochs} & \textbf{acc} \\
\hline
\textit{Mallat} & \textit{XCM} & $8 / 16$ & $300$ & $0.957$ \\
\hline
\textit{UWaveGestureLibraryAll} & \textit{XCM} & $32 / 64$ & $300$ & $0.974$ \\
\hline
\textit{InsectWingbeatSound} & \textit{XCM} & $8 / 16$ & $20$ & $0.865$ \\
\hline
\end{tabular}
}
\caption{Hyperparameters / Results for Binary Classification}
\label{tab:binary_res}
\end{table}

\subsection{Saliency Map Generation for Time Series}
\label{sec:cam}

There are several techniques used in deep learning to visualize areas of an input that are most relevant to predicting a certain class, e.g., \emph{Grad-CAM}
\cite{Selvaraju:2017}, \emph{Grad-CAM++}~\cite{Chattopadhyay:2018}, \emph{Score-CAM}~\cite{Wang:2020},
and \emph{LayerCAM}~\cite{Jiang:2021}. They provide a way to interpret the decision made by an ANN model (with compatible architecture) by highlighting the regions of the input
that contribute the most to the classification result.
The basic idea behind vanilla \emph{Grad-CAM} is to use the gradients of the output class with respect to the feature
maps of the last convolutional layer in the network~\cite{Selvaraju:2017}, i.e., to ask how much the output would change if the corresponding element in the feature map were to change.
The resulting saliency maps can be overlaid on top of the original input to provide a visual explanation.
The regions with higher values in the saliency map correspond to the regions of the input that had the greatest impact on the model's prediction.
The general method of using gradients to visualize the importance of input features is based on the idea of a
hierarchy of feature representations, as in CNNs, where the output of each layer can be interpreted as a set of learned features that capture increasingly complex aspects of the input.
With time series data, the gradient-based approaches work with $1D$ convolutional layers along the temporal dimension of the input.
Each of the methods receives the normalized time series values $V' \in \mathbb{R}^n$, the trained model $M$, and an optional prediction $y$ (default is the
``best guess'', i.e., $y = \operatorname*{argmax}_i P(i \thinspace | \thinspace V') \thinspace \forall \thinspace i \in \{0, 1\}$) as input, and outputs a heatmap $H \in [0, 1]^n$.
Each value $h_i \in H$ rates the importance of a corresponding time series value $v_i \in V', \thinspace \forall \thinspace i \in \{1, \dots n\}$.
Since the XCM~\cite{Fauvel:2021} model allows for relatively straightforward extraction of saliency maps, we use it to demonstrate the principle.
In future work, it might be worthwhile to compare different generation techniques and evaluate their respective advantages and disadvantages for knowledge discovery.

\subsection{Univariate and Multivariate Clustering}
\label{sec:clustering}

\begin{figure}[H]
    \centering
    \includegraphics[width=0.48\textwidth]{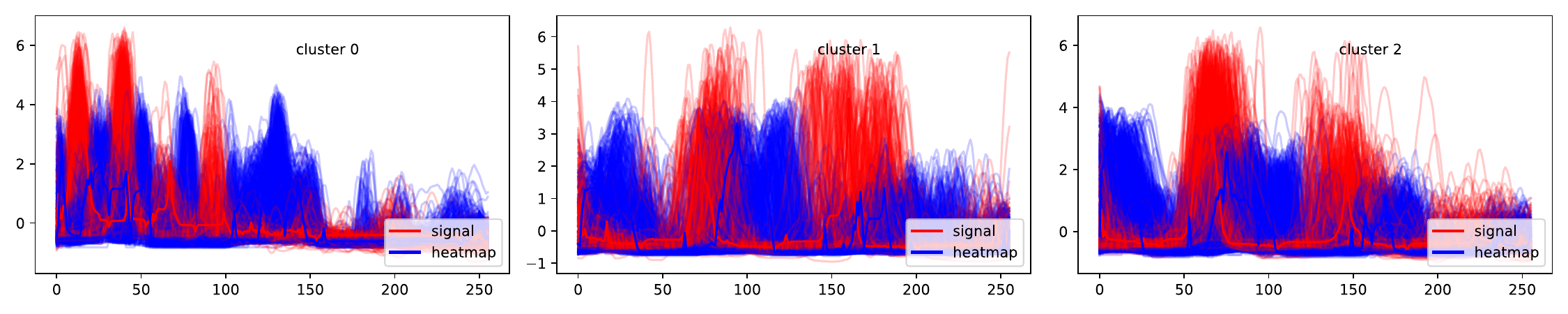}
    \caption[]{Multivariate Clusters \textit{InsectWingbeatSound} (class $0$)}
    \label{fig:multivar_clusters1}
\end{figure}

\begin{figure}[H]
    \centering
    \includegraphics[width=0.48\textwidth]{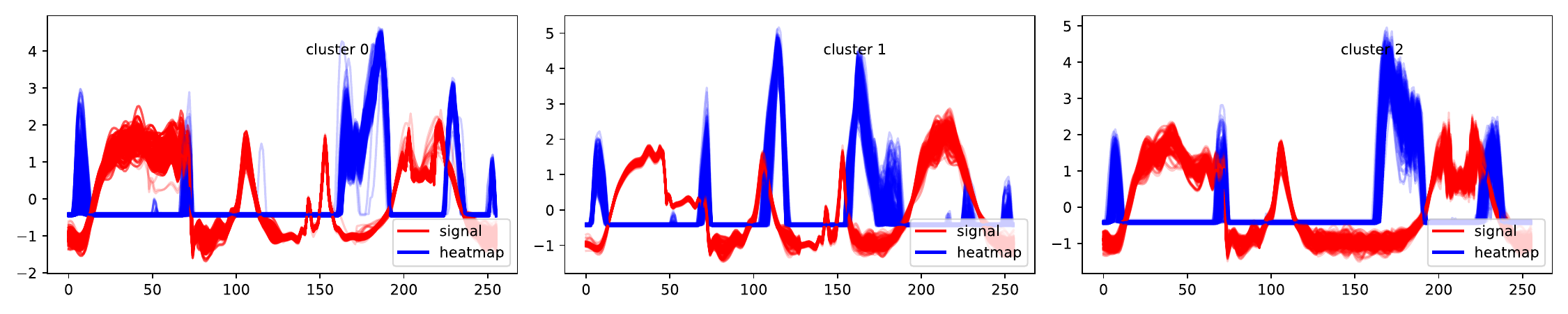}
    \caption[]{Multivariate Clusters \textit{Mallat} (class $0$)}
    \label{fig:multivar_clusters2}
\end{figure}

This section is concerned with the clustering of the input signals, grouped by predicted class ($0$, $1$), resulting from the binary classification, with the objective of subclass identification.
We evaluate three distinct clustering scenarios:
\begin{itemize}
    \item univariate (\textbf{input signals} only)
    \item univariate (\textbf{saliency maps} only)
    \item multivariate (\textbf{input signals + saliency maps})
\end{itemize}
Therefore, in total, there are $18$ different configurations, i.e., clusterings: $3$ datasets $\times$ $2$ classes $\times$ $3$ modes.
The aim is to assess whether the generated saliency maps are beneficial for the clustering process.
Given that all of the employed clustering methods are $k$-means variations, the initial step is to determine a reasonable $k$ value, i.e., the estimated number of subclasses.
Since we are treating it as a knowledge discovery problem, the number of subclasses to be expected is unknown and must be estimated.
This is achieved by using the
\textit{Elbow method} with \textit{Dynamic Time Warping} (DTW) as metric and $k \in [1, 10]$.
After experimenting with different $k$-means approaches and metrics, \textit{Dynamic Time Warping Barycenter Averaging} (DBA) $k$-means
proved to be the most effective.
The $k$ range is set according to the number of classes defined as a criterion in the dataset selection (cf. Sec.~\ref{sec:dateset_sel}), as well as the fact that the two classes are clustered separately.
Setting the upper bound of the $k$ range to an excessively high value can distort the \textit{Elbow} estimation.
To avoid local minima, the DBA $k$-means is always performed $20$ times with different centroid seeds.
The maximum number of iterations is set to $500$ for the $k$-means algorithm and to $300$ for computing the barycenter (centroid) of each cluster.
DTW (used in DBA $k$-means) is not scale-invariant, i.e., differences in scale across variables can dominate the distance calculation.
We are typically more interested in relative shape similarity than absolute magnitude of the time series. Thus, having no or poor normalization can distort the clusters.
In the multivariate case, the first dimension corresponds to the $z$-normalized signals, while the second dimension corresponds to a saliency map for each.
To ensure that each dimension contributes equally, $z$-normalization is also applied to the saliency maps.
Experimentally, this turned out to be an advantage.
To provide two examples, the resulting multivariate clusters for \textit{InsectWingbeatSound} (class $0$) are shown in Fig.~\ref{fig:multivar_clusters1}
and the ones for \textit{Mallat} (class $0$) in Fig.~\ref{fig:multivar_clusters2}.
It is evident that some clusters contain rather uniform samples, while others appear more chaotic.
The saliency maps can be regarded as a relevance guide. Thus, the clustering is based on both the shape (via the signal) and the temporal relevance (via the saliency map).
It is important to note that clustering with input samples in isolation is not merely clustering of input samples, because it uses the information of the binary
predictions -- it is a kind of informed clustering of input samples. Adding the saliency maps as a second channel results in an even more informed clustering.
Another crucial aspect is that the saliency maps are obviously only required in relatively hard to distinguish cases, where additional information is needed.
In datasets with easily distinguishable classes, it may be less beneficial, but such datasets are also not very problematic.

\subsection{LLM-Based Symbolic Approximation and Fuzzy KG Matching}
\label{sec:llm_approx}

The aim of this section is to associate the established cluster centroids (barycenters, i.e., ``average'' time series computed using DBA) with symbolic knowledge.
The identified subclasses are matched to instances in an existing KG to retrieve corresponding semantic information.
This constitutes a fusion of subsymbolic and symbolic representations, where learned patterns in sensor signals are mapped to interpretable concepts from a KG.
It can be regarded as probabilistic semantics, i.e., LLM-based approximation of semantics and subsequent fuzzy matching of these approximations to symbolic properties.
The procedure begins with constructing a KG containing symbolic attributes for each class (cf. Sec.~\ref{sec:symbolic_kr}).
Subsequently, an LLM is provided with the cluster centroids and generates textual features that can be matched to KG entries.
Accordingly, the LLM plays a dual role: (1) translating subsymbolic patterns (e.g., centroids and saliency maps) into symbolic approximations, and (2) performing fuzzy matching against predefined
symbolic properties of the KG. The core pattern discovery occurs at the subsymbolic level.
The LLM serves as a bridge by assigning names and contextual information to cluster centroids, thereby facilitating the actual knowledge discovery.

All experiments indicate that matching is effective using cluster centroids, despite them not being actual samples.
In the multivariate case, saliency maps are likewise represented by centroids. If a centroid is successfully matched, all associated symbolic contextual
information is available. Otherwise, a previously unknown pattern is discovered that should be investigated by a domain expert.
For expert verification, saliency maps can again be very valuable, since they help domain experts to judge whether what the model learned is reasonable.
Essentially, it is a form of supporting evidence.

Without conducting a large-scale comparison of different state-of-the-art LLMs, several \textit{OpenAI} models have been compared (accessed via the \textit{OpenAI} platform API) in terms of their ability of signal description, matching and the
consistency in that task. A very clear winner in that was \textit{o3-2025-04-16}, which, according to the documentation, is \textit{OpenAI}'s ``most powerful reasoning model'' at the time
of writing this paper. It has a context window of $200.000$, a maximum output token number of $100.000$, and it supports both image and text input.
Nonetheless, it is noteworthy that the approach does not depend on this particular LLM model; it may also function with similarly well-suited open-weight models.

\begin{figure}[H]
    \centering
    \includegraphics[width=0.48\textwidth]{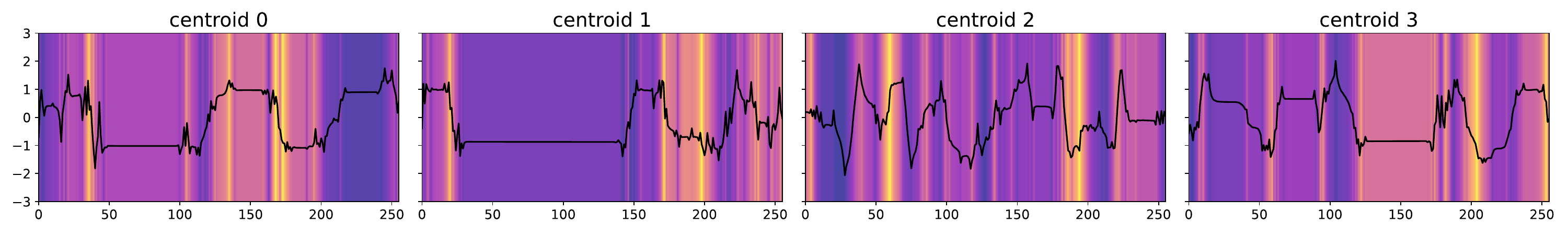}
    \caption[]{Centroids for \textit{UWaveGestureLibraryAll} (Class $1$)}
    \label{fig:centroids4llm_uwave}
\end{figure}

\begin{figure}[H]
    \centering
    \includegraphics[width=0.48\textwidth]{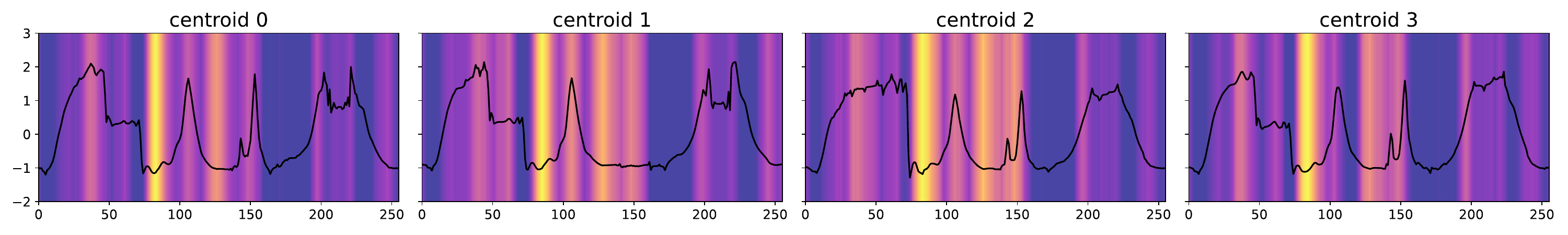}
    \caption[]{Centroids for \textit{Mallat} (Class $1$)}
    \label{fig:centroids4llm_mallat}
\end{figure}

In order to obtain a structured process across the different datasets, the initial textual descriptions of classes that are part of the ontology are also generated with the LLM as
a sort of approximated summarization of the signals.
Obviously, they are not generated based on the clustering results, but based on the initial $z$-normalized dataset. Specifically, each class description is generated by forwarding the
corresponding medoid (real sample that best represents the entire set for each class) to the LLM for description.
Let $S_c$ denote the set of all time series in the dataset that belong to class $c$:
%\vspace{-5pt}
\[
\text{medoid}(S_c) = \arg\min_{x_i \in S_c} \sum_{x_j \in S_c} d(x_i, x_j)
\]
The distance function $d$ represents the pairwise \textit{Dynamic Time Warping} (DTW).
%To give an example, for the \textit{UWaveGestureLibraryAll} dataset, the medoids for each class are shown in Fig.~\ref{fig:medoids}:
% \begin{figure}[H]
%     \centering
%     \includegraphics[width=0.4\textwidth]{img/medoids.pdf}
%     \caption[]{Class Medoids for \textit{UWaveGestureLibraryAll}}
%     \label{fig:medoids}
% \end{figure}
Initially, the symbolic descriptions of each class of a dataset were created manually based on corresponding medoids. However, in order to adopt a more systematic approach without manual biases,
we decided to provide the medoids of a dataset to an LLM that generates the descriptions. The used model is again \textit{o3-2025-04-16}.
This process involved quite a bit of prompt engineering, e.g., deciding on whether it is allowed to include raw values and information about positive and negative parts, or whether it should just
be about the general shape. After conducting several experiments, it turned out to be beneficial to generally incentivize brevity and generality, i.e., to not include raw values.
The performance deteriorated as the descriptions became more nuanced. The LLM focuses extensively on the exact specifics of the medoids then (overfitting).
Two options were implemented for the manner in which the LLM receives the medoids: as \textit{base64}-coded images or as lists of floats. Since the performance was similar in both cases, we opted for the more direct approach of providing medoids as lists of floats.
The exact prompting and extraction of symbolic approximations can be found in the open source repository\footnote{https://github.com/tbohne/saliency\_kd}.
Based on the experiments conducted, the following general conclusions can be drawn: It is impractical to provide the LLM with the entire clusters, as this merely adds input tokens and complexity. The centroids are ideal
representations of the clusters' samples. Additionally, the saliency maps are not beneficial for the matching process. As before, keeping the prompt simple helps, not being overly specific.
We avoid using raw values via prompts and rely on general shape descriptions for performance reasons.
Based on the signal descriptions, the ontology is instantiated for the particular dataset, i.e., corresponding semantic facts are generated.
Afterwards, the LLM analysis takes place, i.e., the centroids (cf. Fig.~\ref{fig:centroids4llm_uwave}, Fig.~\ref{fig:centroids4llm_mallat}) are forwarded to the LLM. On the one hand, for symbolic approximation, and on the other hand,
for potentially matching KG entries.
It is important to note that, in this case, the centroids are multivariate, consisting of the signals themselves and the overlaid saliency maps. Nevertheless, the LLM only receives the signals for description and matching,
since it would be unrealistic to assume saliency map-like features in an existing KG. Practically, it is a lot more plausible to obtain some symbolic descriptions of, e.g., fault curves, and not particular
highlighted segments.
The shape of the final result follows a simple scheme, e.g., $<class\_7, class\_1, class\_3>$. Thus, in this case, the LLM received three centroids and matched
all of them to a particular instance in the KG. Ultimately, additional information for each class can be retrieved from the KG, e.g., the severity of a fault.
In general, such a matching provides valuable context information.
To consider the inherent uncertainty of the LLM-based description and matching and to study the stability of predictions, the final matching process is repeated five times for each dataset ($r_1$ to $r_5$).
The results are summarized in Table~\ref{tab:matching_res}. Each cell indicates the predicted class for each forwarded cluster centroid, and the result column corresponds to a most prominent prediction for each.
The ground truth column indicates the most prevalent ground truth label of the corresponding clusters.
The last column counts the matched selected clusters, i.e., the identified subclasses, and the ground truth classes in brackets.
Matching is only reasonable for the input samples in isolation and the multivariate case, not for the saliency maps in isolation.
\begin{table*}[ht]
\centering
\footnotesize{
\begin{tabular}{|c|c|c|c|c|c|c|c|c|}
\hline
\textbf{mode} & $\bold{r_1}$ & $\bold{r_2}$ & $\bold{r_3}$ & $\bold{r_4}$ & $\bold{r_5}$ & \textbf{result} & \textbf{ground truth} & \textbf{identified} \\
\hline
\textit{UWaveGestureLibraryAll}\_$0$ & $<\_,2,\bold{5},7>$ & $<1,2,8,7>$ & $<\bold{3},2,8,7>$ & $<1,2,\bold{5},7>$ & $<\bold{3},2,\_,7>$ & $<\bold{3},2,\bold{5},7>$ & $<3,7,5,1>$ & $\bold{2/4} (4)$ \\
\hline
\textit{UWaveGestureLibraryAll}\_$1$ & $<5,\_,\bold{2},\bold{8}>$ & $<4,\_,3,5>$ & $<5,\_,3,4>$ & $<5,\bold{4},3,\bold{8}>$ & $<8,\_,3,\_>$ & $<5,\bold{4},3,\bold{8}>$ & $<6,4,2,8>$ & $\bold{2/4} (4)$ \\
\hline
\textit{InsectWingbeatSound}\_$0$ & $<\bold{11},\bold{1},5>$ & $<2,\bold{1},5>$ & $<\bold{11},4,5>$ & $<\bold{11},4,5>$ & $<\bold{11},\bold{1},8>$ & $<\bold{11},\bold{1},5>$ & $<11,1,7>$ & $\bold{2/3} (6)$ \\
\hline
\textit{InsectWingbeatSound}\_$1$ & $<3,2,4>$ & $<3,2,4>$ & $<3,2,4>$ & $<3,2,4>$ & $<3,2,4>$ & $<3,2,4>$ & $<4,10,8>$ & $0/3 (5)$ \\
\hline
\textit{Mallat}\_$0$ & $<1,4,8>$ & $<2,4,3>$ & $<1,4,\bold{7}>$ & $<\bold{5},4,\bold{7}>$ & $<8,\bold{1},\bold{7}>$ & $<1,4,\bold{7}>$ & $<5,1,7>$ & $\bold{1/3} (4)$ \\
\hline
\textit{Mallat}\_$1$ & $<5,\bold{2},3,\bold{4}>$ & $<4,\bold{2},3,\bold{4}>$ & $<\bold{8},1,3,\bold{4}>$ & $<\bold{8},\bold{2},3,\bold{4}>$ & $<2,\bold{2},3,\bold{4}>$ & $<\bold{8},\bold{2},3,\bold{4}>$ & $<8,2,6,4>$ & $\bold{3/4} (4)$ \\
\hline
\textcolor{neuro_color}{\textit{UWaveGestureLibraryAll}\_$0$} & $<\bold{1},2,\_,\_>$ & $<\bold{1},\bold{7},\_,2>$ & $<\bold{1},\bold{7},\_,\_>$ & $<\bold{1},2,\_,\_>$ & $<7,2,8,\_>$ & $<\bold{1},2,8,2>$ & $<1,7,3,5>$ & $1/4 (4)$ \\
\hline
\textcolor{neuro_color}{\textit{UWaveGestureLibraryAll}\_$1$} & $<\bold{4},\bold{8},7>$ & $<\bold{4},\bold{8},7>$ & $<\bold{4},\bold{8},7>$ & $<\bold{4},5,7>$ & $<\bold{4},\bold{8},7>$ & $<\bold{4},\bold{8},7>$ & $<4,8,2>$ & $2/3 (4)$ \\
\hline
\textcolor{neuro_color}{\textit{InsectWingbeatSound}\_$0$} & $<3,4,\bold{11}>$ & $<3,4,\bold{11}>$ & $<3,4,\bold{11}>$ & $<6,4,\bold{11}>$ & $<6,4,\bold{11}>$ & $<3,4,\bold{11}>$ & $<7,1,11>$ & $1/3 (6)$ \\
\hline
\textcolor{neuro_color}{\textit{InsectWingbeatSound}\_$1$} & $<11,\bold{4},2>$ & $<6,\bold{4},2>$ & $<6,\bold{4},2>$ & $<6,\bold{4},2>$ & $<2,\bold{4},2>$ & $<6,\bold{4},2>$ & $<10,4,6>$ & $\bold{1/3} (5)$ \\
\hline
\textcolor{neuro_color}{\textit{Mallat}\_$0$} & $<6,2,4>$ & $<6,2,4>$ & $<1,8,4>$ & $<3,8,4>$ & $<3,8,\bold{1}>$ & $<6,8,4>$ & $<5,7,1>$ & $0/3 (4)$ \\
\hline
\textcolor{neuro_color}{\textit{Mallat}\_$1$} & $<\bold{6},2,\bold{2}>$ & $<\bold{6},2,\bold{2}>$ & $<\bold{6},2,\bold{2}>$ & $<\bold{6},2,\bold{2}>$ & $<\bold{6},2,\bold{2}>$ & $<\bold{6},2,\bold{2}>$ & $<6,8,2>$ & $2/3 (4)$ \\
\hline
\end{tabular}
}
\caption{Accumulated Results for the LLM-Based Matching: multivariate, \textcolor{neuro_color}{input}}
\label{tab:matching_res}
\end{table*}

\section{Evaluation of Knowledge (Re)Discovery}
\label{sec:eval}

This section is concerned with the evaluation of the success of the knowledge (re)discovery process, i.e., the number of identified classes, the number of correctly matched samples, and the comparison
of the saliency map-based variant and the signals-only variant.
A valid initial question is whether it is not possible to simply match the input signals to the KG entries. In the end, the signal dimension of the final centroids is matched, not the saliency maps.
Thus, in principle, it would be possible to only cluster the input signals. However, the clustering is significantly improved by incorporating the saliency maps as a cornerstone for the knowledge discovery.
In other words, it is advantageous to consider not only the signals themselves, but also some information about ROIs (attention over time).
The saliency maps provide additional knowledge, not only the signals themselves, but also where to look at in the signals, which seems to be a crucial information in this context.
Likewise, the saliency maps in isolation only provide information regarding points in time that are deemed important; the multivariate case combines this information with the actual signal.
The expectation should be that the heatmap-aided clustering is significantly superior to the clustering of the input samples alone across multiple datasets.
Finally, for the purpose of evaluation, the most prominent ground truth label in each cluster is determined to be the assumed class.
To demonstrate a successful saliency map-based knowledge discovery, two criteria must be fulfilled:
%\vspace{-3pt}
\begin{enumerate}
    \item The multivariate (heatmap-based) clustering must be significantly superior to the clustering of the signals in isolation.
    \item The identified subclasses (and matched signals) based on the multivariate clustering results must be significantly superior to the ones of the signals alone.
\end{enumerate}
%\vspace{-4pt}
Thus, in general, it has to be demonstrated that saliency maps can play a crucial role in subclass identification.
We begin with a threefold comparison of the clustering performance based on the three scenarios outlined in Sec. \ref{sec:clustering}.
Let $X := \{X_1, X_2, \ldots, X_r\}$ denote the established clustering and $Y := \{Y_1, Y_2, \ldots, Y_s\}$ the ground truth clustering for the total number of samples $n$.
Each entry $n_{ij}$ represents the number of common samples between $X_i$ and $Y_j$, i.e., $n_{ij} = |X_i \cap Y_j|$.
Also, $a_i = \sum_{j=1}^{s} n_{ij}$ and $b_j = \sum_{i=1}^{r} n_{ij}$.

The initially considered metric is the \textit{Adjusted Rand Index} (ARI):
\[
\text{ARI} := \frac{
\sum_{ij} \binom{n_{ij}}{2}
- \frac{
\sum_i \binom{a_i}{2} \sum_j \binom{b_j}{2}
}{
\binom{n}{2}
}
}{
\frac{1}{2} \left[
\sum_i \binom{a_i}{2} + \sum_j \binom{b_j}{2}
\right]
- \frac{
\sum_i \binom{a_i}{2} \sum_j \binom{b_j}{2}
}{
\binom{n}{2}
}
}
\]
The second one is the \textit{Normalized Mutual Information} (NMI), measuring homogeneity and completeness for $Y$ and $X$:
\[
\text{NMI}(Y, X) := \frac{2 \cdot I(Y, X)}{H(Y) + H(X)}
\]
$I$ measures the mutual information using the Shanon entropy $H$:
\[
I(Y, X) := H(Y) - H(Y|X)
\]
Finally, we also consider the purity $P$ of clusters $X$ and classes $Y$, i.e., the extent to which clusters contain a single class:
\[
P := \frac{1}{n} \sum_{X_i \in X} \max_{Y_j \in Y} |X_i \cap Y_j|
\]

The following three tables (Tab.~\ref{tab:cluster_perf_ds1}, Tab.~\ref{tab:cluster_perf_ds2}, Tab.~\ref{tab:cluster_perf_ds3}) quantify the clustering performances for the UCR datasets under consideration.
The $k (gt)$ column refers to both the estimated $k$ using the \textit{Elbow} method and the ground truth number of classes. The final column presents the cluster distribution.\newline
\begin{table}[H]
\centering
\footnotesize{
\begin{tabular}{|c|c|c|c|c|c|}
\hline
\textbf{mode} & $\bold{ARI}$ & $\bold{NMI}$ & $\bold{P}$ & $\bold{k (gt)}$ & \textbf{distribution} \\
\hline
input\_$0$ & $0.426$ & $0.476$ & $0.712$ & $\bold{4 (4)}$ & $\bold{[434, 630, 400, 307]}$ \\
\hline
input\_$1$ & $0.511$ & $0.520$ & $0.700$ & $3 (4)$ & $[591, 672, 500]$ \\
\hline
saliency\_$0$ & $0.106$ & $0.107$ & $0.421$ & $3 (4)$ & $[644, 670, 457]$ \\
\hline
saliency\_$1$ & $0.530$ & $0.516$ & $0.759$ & $\bold{4 (4)}$ & $\bold{[435, 499, 337, 492]}$ \\
\hline
multivariate\_$0$ & $\bold{0.615}$ & $\bold{0.580}$ & $\bold{0.823}$ & $\bold{4 (4)}$ & $\bold{[359, 446, 434, 532]}$ \\
\hline
multivariate\_$1$ & $\bold{0.773}$ & $\bold{0.737}$ & $\bold{0.889}$ & $\bold{4 (4)}$ & $\bold{[384, 471, 451, 457]}$ \\
\hline
\end{tabular}
}
\caption{Clustering Performance \textit{UWaveGestureLibraryAll}}
\label{tab:cluster_perf_ds1}
\end{table}
\vspace{-10pt}
Table~\ref{tab:cluster_perf_ds1} shows that the multivariate mode clearly outperforms the clustering based on input samples alone according to every metric, with the following improvements:
ARI ($c_0$: $44.4\%$, $c_1$: $51.3\%$), NMI ($c_0$: $21.8\%$, $c_1$: $41.7\%$), $P$ ($c_0$: $15.6\%$, $c_1$: $27\%$). $k$ estimation and overall cluster distributions are correspondingly also improved.
Interestingly, for $c_1$, the saliency maps in isolation even lead to a better performance than the input samples in isolation.
Overall, the multivariate case promotes an initial clustering of moderate quality to a satisfactory one.\newline
\begin{table}[H]
\centering
\footnotesize{
\begin{tabular}{|c|c|c|c|c|c|}
\hline
\textbf{mode} & $\bold{ARI}$ & $\bold{NMI}$ & $\bold{P}$ & $\bold{k (gt)}$ & \textbf{distribution} \\
\hline
input\_$0$ & $0.018$ & $0.064$ & $0.233$ & $3 (6)$ & $[526, 331, 193]$ \\
\hline
input\_$1$ & $0.007$ & $0.019$ & $0.226$ & $3 (5)$ & $[325, 245, 360]$ \\
\hline
saliency\_$0$ & $0.165$ & $0.274$ & $0.358$ & $3 (6)$ & $[272, 525, 253]$ \\
\hline
saliency\_$1$ & $0.118$ & $0.187$ & $0.309$ & $3 (5)$ & $[352, 309, 269]$ \\
\hline
multivariate\_$0$ & $\bold{0.233}$ & $\bold{0.368}$ & $\bold{0.398}$ & $3 (6)$ & $[307, 281, 462]$ \\
\hline
multivariate\_$1$ & $\bold{0.169}$ & $\bold{0.248}$ & $\bold{0.326}$ & $3 (5)$ & $[254, 303, 373]$ \\
\hline
\end{tabular}
}
\caption{Clustering Performance \textit{InsectWingbeatSound}}
\label{tab:cluster_perf_ds2}
\end{table}
\vspace{-10pt}
Table~\ref{tab:cluster_perf_ds2} confirms that the multivariate mode significantly outperforms the clustering based on input samples alone.
Once more, in accordance with every metric, with the following improvements:
ARI ($c_0$: $1194.4\%$, $c_1$: $2314.3\%$), NMI ($c_0$: $475\%$, $c_1$: $1205.3\%$), $P$ ($c_0$: $70.8\%$, $c_1$: $44.2\%$).
For this dataset, pure saliency maps even lead to a significantly better performance than pure input samples in every case.
In the end, the saliency map-guided clustering transforms an infeasibly poor clustering into a fairly moderate one.
Once again, it is not about absolute results, which can be further optimized in future work; it is about the improvement, which is substantial and consistent.
The relative improvement is exceptionally high in this case due to the dataset being challenging to classify and the baseline yielding unsatisfactory results.\newline
\begin{table}[H]
\centering
\footnotesize{
\begin{tabular}{|c|c|c|c|c|c|}
\hline
\textbf{mode} & $\bold{ARI}$ & $\bold{NMI}$ & $\bold{P}$ & $\bold{k (gt)}$ & \textbf{distribution} \\
\hline
input\_$0$ & $\bold{0.651}$ & $\bold{0.807}$ & $\bold{0.713}$ & $3 (4)$ & $[608, 290, 251]$ \\
\hline
input\_$1$ & $0.634$ & $0.803$ & $0.721$ & $3 (4)$ & $[278, 625, 293]$ \\
\hline
saliency\_$0$ & $0.585$ & $0.681$ & $0.702$ & $3 (4)$ & $[290, 536, 323]$ \\
\hline
saliency\_$1$ & $0.901$ & $0.909$ & $0.934$ & $\bold{4 (4)}$ & $\bold{[269, 338, 292, 297]}$ \\
\hline
multivariate\_$0$ & $\bold{0.649}$ & $0.804$ & $\bold{0.713}$ & $3 (4)$ & $[610, 249, 290]$ \\
\hline
multivariate\_$1$ & $\bold{0.918}$ & $\bold{0.931}$ & $\bold{0.940}$ & $\bold{4 (4)}$ & $\bold{[300, 293, 278, 325]}$ \\
\hline
\end{tabular}
}
\caption{Clustering Performance \textit{Mallat}}
\label{tab:cluster_perf_ds3}
\end{table}
\vspace{-10pt}
Finally, Table~\ref{tab:cluster_perf_ds3} is a further compelling case, as it is a dataset that is already quite well-solved with the baseline approach.
Consequently, there is limited potential for enhancement. First, for $c_0$, all metrics are equal to negligible deviations in both the input-only and multivariate cases.
Thus, there is no improvement for $c_0$. This can be partially attributed to the fact that the dataset is a priori quite well-solved.
In certain instances, the inherent structure may render additional information redundant, thereby diminishing the contribution effect.
However, there is a substantial and consistent enhancement for $c_1$ across all metrics:
ARI ($c_1$: $44.8\%$), NMI ($c_1$: $15.9\%$), $P$ ($c_1$: $30.4\%$). Additionally, the $k$ estimation is enhanced and aligns with the ground truth, as well as the resulting cluster distribution.
In this case, the saliency maps appear highly advantageous, which is reinforced by the observation that the saliency maps in isolation significantly outperform the signals in isolation for $c_1$.
Overall, we once again observe a very clear and consistent contribution of the saliency maps, transforming an initially adequate clustering into an almost perfect one.
Ultimately, it was confirmed that the multivariate clustering significantly outperforms the clustering of the signals in isolation across all considered datasets $(1)$, i.e., the saliency maps do have a positive impact.
Despite their occasional utility in isolation, the saliency maps, i.e., ROIs, only truly contribute in conjunction with the signals as context.

Another aspect to be quantified are the subclass predictions introduced in section~\ref{sec:llm_approx}.
The last column in Table~\ref{tab:matching_res} counts the matched selected clusters, i.e., how many classes are discovered, and the ground truth classes in brackets.
A class is considered discovered if the dominant ground truth class within the cluster corresponds to the LLM prediction.
As can be seen in the last column, the saliency map-aided approach wins in five out of six modes. Thus, it not only improves the clustering results, but those improvements also consistently lead to
increased numbers of finally identified subclasses. The sole exception to this is the case of \textit{InsectWingbeatSound\_1}, in which the multivariate case does not result in the identification of any subclasses.
This case led to clusters that were very stably predicted wrong by the LLM. This one exception aligns perfectly with the metrics displayed in Table~\ref{tab:cluster_perf_ds2}.
The multivariate approach significantly improves the clusterings; nonetheless, it remains the most challenging dataset, which also manifests in the final LLM matching.
Ultimately, on a more fine-grained level, we are also interested in the fraction of ground truth samples of a class that are matched to a correctly predicted class for each dataset, i.e., the precision
(proportion of dominant class in cluster in brackets):
\begin{itemize}
  \item \textit{UWaveGestureLibraryAll} (\textbf{input}):\newline $\bold{c_0}$: $\{1: \frac{321}{437} (\frac{321}{434})\}$, $\bold{c_1}$: $\{4: \frac{377}{450} (\frac{377}{591}), 8: \frac{418}{460} (\frac{418}{672})\}$
  \item \textit{UWaveGestureLibraryAll} (\textbf{multivariate}):\newline $\bold{c_0}$: $\{3: \frac{304}{454} (\frac{359}{304}), 5: \frac{385}{433} (\frac{385}{434})\}$, $\bold{c_1}$: $\{4: \frac{367}{450} (\frac{367}{471}), 8: \frac{431}{460} (\frac{431}{457})\}$
  \item \textit{InsectWingbeatSound} (\textbf{input}):\newline $\bold{c_0}$: $\{11: \frac{66}{180} (\frac{66}{193})\}$, $\bold{c_1}$: $\{4: \frac{58}{180} (\frac{58}{245})\}$
  \item \textit{InsectWingbeatSound} (\textbf{multivariate}):\newline $\bold{c_0}$: $\{11: \frac{152}{180} (\frac{152}{307}), 1: \frac{121}{180} (\frac{121}{281})\}$, $\bold{c_1}$: $\emptyset$
  \item \textit{Mallat} (\textbf{input}):\newline $\bold{c_0}$: $\emptyset$, $\bold{c_1}$: $\{6: \frac{278}{294} (\frac{278}{278}), 2: \frac{292}{292} (\frac{292}{293})\}$
  \item \textit{Mallat} (\textbf{multivariate}):\newline $\bold{c_0}$: $\{7: \frac{290}{291} (\frac{290}{290})\}$, $\bold{c_1}$: $\{8: \frac{292}{293} (\frac{292}{300}), 2: \frac{292}{292} (\frac{292}{293}), 4: \frac{262}{289} (\frac{262}{325})\}$
\end{itemize}
As demonstrated in this analysis, the multivariate scenario is not only discovering more subclasses, but also consistently leading to better precision in cases where both scenarios find the subclass, e.g.,
subclass $11$ in the \textit{InsectWingbeatSound} dataset with more than twice as many matched samples. The mean sample coverage across all identified subclasses and datasets is $0.73$ for the input-only case and
$0.87$ for the multivariate case. Consequently, success criterion $(2)$ is also satisfied.

Finally, for practical purposes, one might be interested in evaluations without ground truth labels, e.g., for initial filtering.
Without knowing ground truth labels of signals allocated to a given cluster, we need alternative (uninformed) quantitative evaluation metrics to estimate the quality of a cluster $X_j \in X$ with centroid $C_j$:
%Hence, it is important to distinguish between quantitative evaluations with and without ground truth, since they serve different purposes.
%The one with ground truth information (cf. above) is used to demonstrate the superiority of saliency map-aided clustering.
%The one without, i.e., the one below, is used to demonstrate the actual process in practice.
% cluster variance across samples (for each time step and variable: how much do samples vary? -- cluster spread)
% cluster variance across time steps (for each sample and variable: how much does the sample fluctuate over time?)
% cluster variance across variables (for each sample and time step: how different are the variables at that moment?)
%\textcolor{red}{We consider the intra-cluster DTW distances $\mathrm{DTW}_{\mathrm{intra}}$,
%inter-cluster DTW distances $\mathrm{DTW}_{\mathrm{inter}}$,
%DTW silhouette score $S_{\mathrm{DTW}}$,
%cluster size entropy $H_{\mathrm{cluster}}$,
%total variance $\sigma^2_{\mathrm{total}}$,
%cluster variance across samples $\sigma^2_{\mathrm{samples}}$,
%cluster variance across time steps $\sigma^2_{\mathrm{time}}$,
%cluster variance across variables $\sigma^2_{\mathrm{vars}}$,
%intra-class variance $\sigma^2_{\mathrm{intra}}$,
%and the average cluster size $\mathrm{avg}(n_k)$.}
\begin{itemize}
    \item $\mathrm{DTW}_{\mathrm{intra}} := \frac{1}{n_j} \sum_{i=1}^{n_j} d(x^j_i, C_j), X_j \in X, x^j_i \in X_j$
    \item $\mathrm{DTW}_{\mathrm{inter}} := \frac{1}{r - 1} \sum_{\substack{k=1, k \neq j}}^{r} d(C_j, C_k)$ for $X_j \in X$
    \item $\mathrm{DTW}_{\mathrm{frac}} := \frac{\mathrm{DTW}_{\mathrm{intra}}}{\mathrm{DTW}_{\mathrm{inter}}}$
    \item $S_{\mathrm{DTW}}$: \textit{Silhouette} score based on DTW distance
    \item cluster size entropy $\eta := -\frac{1}{\log r} \sum_{j=1}^r \frac{n_j}{N} \log \frac{n_j}{N}, N = \sum_{j=1}^r n_j$
    \item $\sigma^2_\mathrm{s} := \frac{1}{n_j T} \sum_{i=1}^{n_j} \sum_{t=1}^T (x_{it}^j - \frac{1}{n_j} \sum_{k=1}^{n_j} x_{kt}^j)^2, T = |x_i^j|, x_i^j \in X_j$
    \item $\sigma^2_\mathrm{i} := \frac{1}{n_j T} \sum_{i=1}^{n_j} \sum_{t=1}^T (x_{it}^j - c_t^j)^2, T = |x_i^j|, x_i^j \in X_j, c_t^j \in C_j$
\end{itemize}
With the exception of $\eta$ and $S_{\mathrm{DTW}}$, all metrics can be applied to either the individual clusters or the overall clustering outcome via averaging.
We are interested in whether these uninformed metrics can serve as indicators of ground truth performance and therefore calculate the correlation between the two.
There are medium to strong correlations with NMI for all metrics except for $\eta$. The strongest correlation ($\rho = -0.81$) is observed with $\sigma^2_\mathrm{s}$.
Further medium to strong correlations with NMI exist for $\sigma^2_\mathrm{i} (\rho = -0.77)$, $S_{\mathrm{DTW}} (\rho = 0.72)$ and $\mathrm{DTW}_{\mathrm{frac}} (\rho = -0.72)$,
with corresponding $p$-values $< 0.01$.
Medium correlations of these metrics to ARI and $P$ can also be found; however, these are slightly weaker, and partially insignificant.
Therefore, in general, the metrics do correlate with the ground truth evaluations and are as such good predictors.
However, an examination of the correlations between these metrics and the extent to which the centroid of a cluster is correctly matched by the LLM suggests the need for more sophisticated filtering algorithms,
e.g., the accumulation of multiple metrics for the purpose of cluster evaluation.

\section{Conclusions}
\label{sec:conclusions}

Experimental results on well-established time series classification datasets demonstrate the effectiveness of our saliency map-driven method for knowledge discovery, significantly outperforming signal-only baselines in both
clustering and subclass identification. In Sec. \ref{sec:eval}, we define two criteria for success, both of which are satisfied. It has been confirmed that the multivariate heatmap-aided clustering is significantly
superior to the clustering of the signals in isolation across all considered datasets $(1)$, and that these improvements also consistently lead to substantially increased numbers of finally identified subclasses and
matched signals $(2)$.
The pivotal role of the saliency maps lies in enhancing the quality of clustering, i.e., actually discriminating classes that are hardly distinguishable in shape.
This, in turn, enhances the efficacy of the final LLM-based matching due to more precise cluster centroids that better reflect the underlying classes.
Thus, we proved our initial hypothesis of saliency maps having a beneficial effect in subclass discovery -- despite the fact that state-of-the-art saliency map generation techniques are suboptimal and have certain weaknesses
and problems with regard to consistency and robustness when applied to time series data, as argued in~\cite{Balestra:2023}.
The precise semantics and dynamics of time series are not utilized to their full potential by current saliency map generation techniques.
Consequently, enhancements to saliency map generation techniques could also lead to further enhancements in saliency map-guided knowledge discovery.

There are several natural next steps, e.g., investigating the selection of the number of classes in label subsumption based on a preliminary analysis of several groups, i.e., fewer than the original
number, but potentially more than two.
In addition, it could be reasonable to separate hard-to-distinguish subclasses during label subsumption, thereby forcing the model to learn the nuances.
If the separation of the two classes is too straightforward, it may result in non-accurate clusters due to the fact that tiny features are simply not required to be learned.
Moreover, it might be worthwhile to compare different saliency map generation techniques and evaluate their respective advantages and disadvantages in terms of knowledge discovery.
In future work, it would also be reasonable to investigate other methods of signal summarization, such as characteristic feature extraction via FFT, statistical features, etc.
Also, the LLM-based matching could be replaced by fuzzy text matching, cosine similarity over embeddings, or even rule-based matchings.
Another extension could be the filtering of established clusters. Not all established clusters must be forwarded to the LLM for symbolic approximation; only those that satisfy certain quantifiable requirements.
A filtering algorithm could preselect suited clusters based on quantifying metrics without ground truth labels.

\begin{acks}
\vspace{-0.25\baselineskip}
%\footnotesize
The \emph{DFKI Niedersachsen} is sponsored by the \emph{Ministry of Science and Culture of Lower Saxony} and the \emph{Volkswagenstiftung}.
The paper describes work carried out in the context of the funded project \emph{RAILX} (BMWK, FKZ: 13IPC039B).
\end{acks}

\bibliographystyle{ACM-Reference-Format}
\bibliography{bibliography}

\end{document}